\pdfoutput=1
\documentclass[11pt]{article}

\usepackage[final]{acl}

\usepackage{times}
\usepackage{latexsym}

\usepackage[T1]{fontenc}
\usepackage[utf8]{inputenc}
\usepackage{microtype}
\usepackage{inconsolata}
\usepackage{graphicx}
\usepackage{booktabs}
\usepackage{multirow}
\usepackage{amsmath}
\usepackage{xcolor}
\usepackage{enumitem}
\usepackage{subcaption}

\usepackage{listings}
\usepackage{amssymb}
\usepackage{pgfplots}
\pgfplotsset{compat=1.18}
\usepackage{pifont}
\usepackage{makecell}
\newcommand{\cmark}{{\color{green!70!black}\ding{51}}}
\newcommand{\xmark}{{\color{red!80!black}\ding{55}}}

\title{Do Vision--Language Models Understand Human Engagement in Games?}

\author{
  Ziyi Wang$^*$ \\
  Texas A\&M University \\
  \texttt{ziyiwang@tamu.edu} \\\And
  Qizan Guo$^*$ \\
  University of Southern California \\
  \texttt{qizanguo@usc.edu} \\\AND
  Rishitosh Kumar Singh$^*$ \\
  Arizona State University \\
  \texttt{rksing18@asu.edu} \\\And
  Xiyang Hu \\
  Arizona State University \\
  \texttt{xiyanghu@asu.edu} \\
}

\begin{document}
\maketitle
{\def\thefootnote{*}\footnotetext{Equal contribution.}}

\begin{abstract}
Inferring human engagement from gameplay video is important for game design and player-experience research, yet it remains unclear whether vision--language models (VLMs) can infer such latent psychological states from visual cues alone.
Using the GameVibe Few-Shot dataset across nine first-person shooter games, we evaluate three VLMs under six prompting strategies.
We consider both pointwise engagement prediction and pairwise prediction of engagement change between consecutive windows.
Results show that zero-shot VLM predictions are generally weak and often fail to outperform per-game majority-class baselines.
Memory- or retrieval-augmented prompting improves pointwise prediction in some settings, whereas pairwise prediction remains consistently difficult across strategies.
Theory-guided prompting alone does not reliably help and can instead reinforce surface-level shortcuts.
These findings suggest a perception--understanding gap in current VLMs: although they can recognize visible gameplay cues, they struggle to robustly infer human engagement across games.
\end{abstract}

\section{Introduction}

Player engagement is a central construct in game design, live streaming, esports analytics, and interactive content creation \cite{boyle2012engagement}.
Understanding when and why players become engaged---or disengaged---has direct implications for adaptive game design, automated highlight detection, and player retention modeling.

Recent VLMs \cite{openai2023gpt4,gemini2023,bai2023qwenvl} have demonstrated strong visual perception, but these capabilities do not always translate into robust task performance \cite{fu2025hiddenplainsightvlms,zhang2025videogamebenchvisionlanguagemodelscomplete}.
A fundamentally different question is whether VLMs can infer how a human feels about what they see---a psychological state requiring contextual reasoning about challenge, progress, and narrative tension.

Games provide an ideal testbed because they decouple visual intensity from engagement.
A scoreboard showing a close match requires multi-step reasoning (recognizing the score $\to$ understanding competitiveness $\to$ inferring tension), while an explosion-filled frame may look intense but correspond to a routine, low-engagement moment.

In this paper, we ask: \textbf{Do current VLMs understand human engagement in games?}
Specifically, we investigate three research questions:
\begin{itemize}[nosep,leftmargin=*]
    \item \textbf{Q1:} Can VLMs predict engagement from gameplay frames, and do few-shot demonstrations or retrieval strategies improve prediction?
    \item \textbf{Q2:} Do theory-aligned probes (Flow, GameFlow, SDT, MDA; Section~\ref{sec:theories}) help VLMs capture engagement-related dimensions?
    \item \textbf{Q3:} Are VLM predictions consistent across different games and across different models?
\end{itemize}

Despite growing interest in VLM-based affect recognition, no prior work has systematically evaluated whether VLMs can predict human engagement from gameplay video, nor investigated how prompting strategies---including theory-grounded probing and cross-game retrieval---affect this capability (Table~\ref{tab:comparison}).
We make the following contributions:
\begin{itemize}[nosep,leftmargin=*]
    \item \textbf{Benchmark.} We present the first systematic evaluation of VLMs on engagement prediction in gameplay videos, testing three architecturally distinct VLMs across 9 games under six prompting strategies of increasing complexity.
    \item \textbf{Methodology.} We introduce theory-guided prompting grounded in four engagement frameworks (Flow, GameFlow, SDT, MDA) and cross-game retrieval-augmented strategies using both VLM-derived and CLIP embeddings, enabling structured comparison of how prompting complexity affects affective inference.
    \item \textbf{Findings.} We identify a fundamental perception--understanding gap: VLMs achieve only $\sim$57\% zero-shot accuracy (below majority-class baselines), conflate visual intensity with engagement, rely on surface-feature shortcuts, and generate confidently wrong rationales. Theory guidance alone amplifies biases rather than compensating for them.
\end{itemize}
We release all code, prompts, and evaluation scripts to facilitate reproducibility and future research.\footnote{\url{https://github.com/Zoewang0213/vlm-game-engagement}}

\section{Background \& Related Work}
\label{sec:related}

Table~\ref{tab:comparison} positions our work relative to the most related prior studies.
Existing VLM evaluations either target non-game affect recognition without game-specific contexts \cite{lu2024gptpsychologist,bhattacharyya2025evaluatingvisionlanguagemodelsemotion,liu2025mmaffbenmultilingualmultimodalaffective}, or evaluate VLMs in game domains on perceptual tasks (action recognition, game completion) rather than affective inference \cite{zhang2025videogamebenchvisionlanguagemodelscomplete,paglieri2024balrog}.
Conversely, game engagement prediction work has relied on supervised models without evaluating VLMs \cite{pinitas2024acrossgame}, or used text-only LLMs without visual input \cite{melhart2025gamevibe}.
Our work is the first to combine all seven dimensions: visual input, VLM evaluation, game domain, engagement prediction, theory-grounded probes, cross-game transfer, and systematic failure analysis.

\begin{table*}[t]
\centering
\caption{Comparison of our work with the most related prior studies across key evaluation dimensions.}
\label{tab:comparison}
\footnotesize
\renewcommand{\arraystretch}{1.2}
\begin{tabular}{@{}lccccccc@{}}
\toprule
\textbf{Study} & \makecell{\textbf{Visual}\\\textbf{Input}} & \makecell{\textbf{VLM}\\\textbf{Evaluation}} & \makecell{\textbf{Game}\\\textbf{Domain}} & \makecell{\textbf{Engagement}\\\textbf{Task}} & \makecell{\textbf{Theory}\\\textbf{Probes}} & \makecell{\textbf{Cross-Game}\\\textbf{Transfer}} & \makecell{\textbf{Failure}\\\textbf{Analysis}} \\
\midrule
\citet{melhart2025gamevibe} & \xmark & \xmark & \cmark & \cmark & \xmark & \cmark & \xmark \\
\citet{bhattacharyya2025evaluatingvisionlanguagemodelsemotion} & \cmark & \cmark & \xmark & \xmark & \xmark & \xmark & \cmark \\
\citet{lu2024gptpsychologist} & \cmark & \cmark & \xmark & \xmark & \xmark & \xmark & \xmark \\
\citet{pinitas2024acrossgame} & \cmark & \xmark & \cmark & \cmark & \xmark & \cmark & \xmark \\
\citet{zhang2025videogamebenchvisionlanguagemodelscomplete} & \cmark & \cmark & \cmark & \xmark & \xmark & \xmark & \xmark \\
\citet{paglieri2024balrog} & \cmark & \cmark & \cmark & \xmark & \xmark & \xmark & \xmark \\
\citet{liu2025mmaffbenmultilingualmultimodalaffective} & \cmark & \cmark & \xmark & \xmark & \xmark & \xmark & \xmark \\
\textbf{Ours} & \cmark & \cmark & \cmark & \cmark & \cmark & \cmark & \cmark \\
\bottomrule
\end{tabular}
\end{table*}

\subsection{Engagement in Games}

Engagement is a multi-dimensional, subjective, and temporally dynamic quality of the user experience \cite{obrien2008user,brockmyer2009development}: unlike objective visual properties, it is a latent psychological state that must be inferred rather than observed.
In games, engagement manifests as a high-arousal, positive-valence state \cite{juvrud2022game}, and multiple studies have modeled it at the moment-to-moment level \cite{sweetser2005gameflow,melhart2020moment,barthet2024gamevibe,pinitas2023predicting}; see \citet{yannakakis2023affectivegame} for a comprehensive survey.

\subsection{LLMs and VLMs for Affect Modelling}

\paragraph{From text to vision.}
LLM-based affective computing has focused primarily on text-based sentiment analysis \cite{zhang2024sentiment,zhang2026affective}.
Multimodal affect modelling from visual input remains largely unexplored: existing work relies on still images \cite{lian2024gpt4v,yang2024emollm,lu2024gptpsychologist,dai2026bikeability} or evaluates whole scenes rather than time-continuous labelling \cite{humanomni2025}.

\paragraph{VLM affect evaluation.}
VLMs consistently lag behind supervised models on affect recognition and exhibit systematic biases \cite{lu2024gptpsychologist,bhattacharyya2025evaluatingvisionlanguagemodelsemotion,liu2025mmaffbenmultilingualmultimodalaffective}, though careful prompt engineering can partially close this gap \cite{liu2024contextualemotion}---motivating our comparison of six prompting strategies.

\paragraph{Fundamental limitations.}
Pre-trained LLMs struggle with in-context learning for affect prediction, remaining anchored to knowledge priors \cite{chochlakis2024strong}, and exhibit inherent biases on subjective tasks \cite{mao2023biases,amin2024wide}.
Our work extends \citet{melhart2025gamevibe}'s evaluation paradigm by introducing theory-grounded probes, cross-game few-shot transfer with multiple embedding strategies, and systematic failure-mode analysis.

\subsection{Theoretical Frameworks as Probes}
\label{sec:theories}

We use four established engagement theories not as assumptions about engagement, but as structured probing tools to test whether VLMs can extract relevant latent dimensions.
These four frameworks are the canonical accounts of engagement in the game affect literature \cite{yannakakis2023affectivegame}, each contributing a distinct dimension absent from the others.
Flow Theory \cite{csikszentmihalyi1990flow} captures the challenge--skill balance; GameFlow \cite{sweetser2005gameflow} adapts it for games with dimensions of control, immersion, and social interaction; Self-Determination Theory (SDT) \cite{ryan2000sdt} targets competence, autonomy, and relatedness; and the Mechanics--Dynamics--Aesthetics (MDA) framework \cite{hunicke2004mda} decomposes the experience into mechanics, dynamics, and aesthetics.
Each theory yields concrete visual indicators that we operationalize as prompt dimensions in our theory-guided strategies (Section~\ref{sec:theory_prompt}).

\subsection{VLMs and Video Understanding}

VLMs excel at perceptual tasks such as VQA \cite{antol2015vqa} and action recognition \cite{simonyan2014twostream}, yet whether they can perform affective reasoning---inferring how a human would feel about a visual scene---remains largely untested \cite{choi2023socket,kamath2023whatsup}.

\paragraph{VLMs in game domains.}
Recent benchmarks consistently reveal VLM limitations in games: near-zero game completion rates \cite{zhang2025videogamebenchvisionlanguagemodelscomplete}, near-random performance from screenshots \cite{hu2025lmgamebenchgoodllmsplaying}, and sometimes worse performance with visual input than text-only descriptions \cite{paglieri2024balrog}.
Our work shows the problem is even more severe when the task requires affective inference beyond perception (Table~\ref{tab:comparison}).

\section{Experimental Setup}
\label{sec:setup}

\begin{figure*}[t]
\centering
\includegraphics[width=\textwidth]{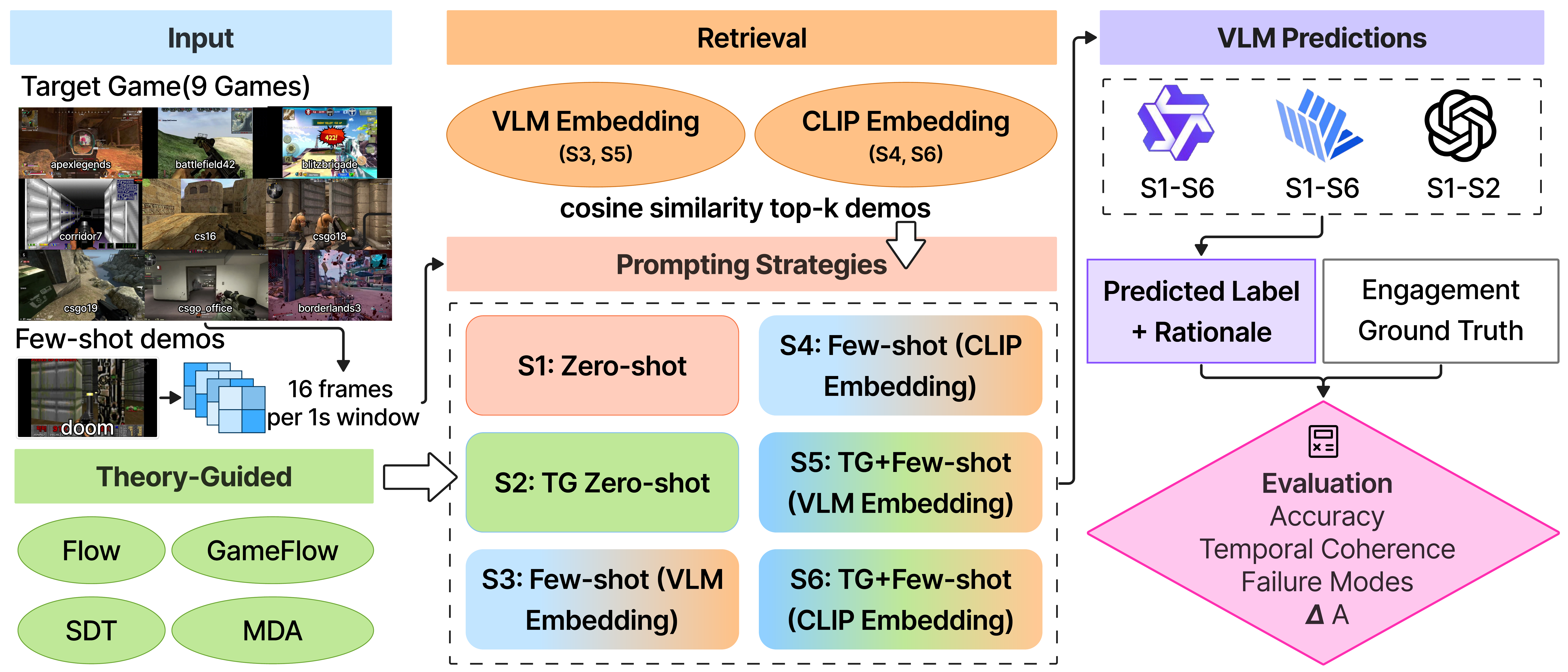}
\caption{Overview of our experimental pipeline. The GameVibe Few-Shot (GVFS) dataset provides 9 target games, each containing 61 seconds of gameplay sampled at 16 frames per second (59 evaluation windows after reaction-lag shifting). We apply six prompting strategies of increasing complexity (S1--S6). Three VLMs---InternVL3.5-8B-Instruct, Qwen3-VL-8B-Instruct, and GPT-4o---predict binary engagement labels (High/Low). Predictions are evaluated against human annotations using accuracy, $\Delta A$ (Eq.~\ref{eq:delta_a}), temporal coherence, and failure-mode analysis.}
\label{fig:pipeline}
\end{figure*}

\subsection{Task Definition}
\label{sec:task_def}

Given a sequence of $n$ frames $\{f_1, \ldots, f_n\}$ sampled from a 1-second gameplay window, the task is to predict the corresponding binary engagement label $y \in \{\text{High}, \text{Low}\}$, where the ground truth is derived from aggregated human annotations.
We evaluate two complementary settings: \textit{pointwise prediction}, where the model predicts the engagement level of each window independently, and \textit{pairwise prediction}, where the model compares two consecutive windows and predicts whether engagement increased or decreased.
This task is fundamentally different from standard visual recognition: engagement is a latent psychological state that cannot be directly observed, requiring the model to reason about challenge, progress, and narrative context beyond surface-level visual features.

\subsection{Dataset and Labels}

We use the public \textit{GameVibe Few-Shot (GVFS)} dataset \cite{barthet2024gamevibe}, derived from the GameVibe corpus of 120 audio-visual clips across 30 FPS games.
The engagement labels are third-party \textit{observer} annotations: each clip is annotated by 5 observers from a pool of 20 trained, gaming-familiar annotators, who provide continuous engagement traces via the RankTrace interface \cite{lopes2017ranktrace} on the PAGAN platform \cite{melhart2019pagan} and were instructed to treat high engagement as a feeling of tension, excitement, and readiness, and low engagement as boredom, low interest, and disassociation \cite{barthet2024gamevibe}.
Following the GameVibe processing pipeline, raw traces are interpolated into 1-second windows and normalized to $[0,1]$ before aggregation.

\paragraph{Temporal windows and frame sampling.}
Following the preprocessing protocol of \citet{pinitas2024acrossgame}, validated for this dataset, we split each 61-second clip into non-overlapping 1-second windows and, to account for human reaction lag, shift the input by 1 second relative to the annotations.
Because the annotation traces cover 60 seconds while each clip spans 61 seconds, the shifted pairing (video window $t$ with annotation $t{+}1$) yields 59 valid evaluation windows per clip.
From each 1-second window we sample 16 frames (i.e., 16 fps), giving 944 frames per game clip.
All frames within a window share the same engagement label; the evaluation unit throughout this paper is the 1-second window.

\paragraph{Annotation aggregation and binarization.}
For each window, we apply trimmed-mean aggregation: discard the maximum and minimum of the five scores and average the remaining three, yielding $e \in [0,1]$.
We binarize: $e \le 0.5 \Rightarrow$ Low, $e > 0.5 \Rightarrow$ High.

\paragraph{Pairwise sample construction.}
For pairwise experiments, consecutive windows with engagement change $|\Delta e| > 0.05$ are labeled Increase or Decrease; near-stable pairs are excluded.
The 0.05 threshold follows \citet{pinitas2024acrossgame}, who use it to isolate meaningful engagement changes from annotation noise.

\paragraph{Game subcorpora.}
We evaluate across 9 games spanning diverse visual styles and pacing (Table~\ref{tab:dataset}), with DOOM as the source domain for cross-game few-shot demonstrations.

\begin{table}[t]
\centering
\small
\begin{tabular}{lccc}
\toprule
\textbf{Game} & \textbf{Windows} & \textbf{High (\%)} & \textbf{Low (\%)} \\
\midrule
Borderlands 3 & 59 & 81.4 & 18.6 \\
CS:GO Office & 59 & 33.9 & 66.1 \\
Blitz Brigade & 59 & 64.4 & 35.6 \\
Corridor 7 & 59 & 39.0 & 61.0 \\
Battlefield 42 & 59 & 39.0 & 61.0 \\
Apex Legends & 59 & 55.9 & 44.1 \\
CSGO19 & 59 & 59.3 & 40.7 \\
CSGO18 & 59 & 22.0 & 78.0 \\
CS 1.6 & 59 & 78.0 & 22.0 \\
\midrule
DOOM (source) & 59 & 45.8 & 54.2 \\
\bottomrule
\end{tabular}
\caption{Label distribution across the 9 evaluation games and the source domain. Each game contributes 61 seconds of gameplay (59 evaluation windows after reaction-lag shifting, 944 frames at 16 fps). All games exhibit varying degrees of class imbalance.}
\label{tab:dataset}
\end{table}

\subsection{Evaluation Protocol}
\label{sec:protocol}

\paragraph{Cross-game few-shot setting.}
Few-shot demonstrations are drawn from DOOM gameplay, while predictions are made on the 9 target games in Table~\ref{tab:dataset}---testing generalization across visually distinct titles.
No model fine-tuning is performed.

\paragraph{VLMs tested.}
We evaluate three VLMs without fine-tuning: \textbf{InternVL3.5-8B-Instruct} \cite{wang2025internvl} and \textbf{Qwen3-VL-8B-Instruct} \cite{qwen2025qwen3} (open-source, all strategies S1--S6), and \textbf{GPT-4o} \cite{openai2024gpt4o} (proprietary, S1--S2 only due to API cost).
Using three architecturally distinct VLMs addresses the concern that observed failures might be model-specific.

\paragraph{Metrics.}
We report raw accuracy and accuracy relative to the per-game majority-class baseline:
\begin{equation}
\Delta A = \frac{A_{\text{model}} - A_{\text{baseline}}}{A_{\text{baseline}}}
\label{eq:delta_a}
\end{equation}
where $A_{\text{model}}$ is the VLM's accuracy and $A_{\text{baseline}}$ is the per-game majority-class accuracy. Positive $\Delta A$ indicates performance above the trivial classifier.
We additionally measure temporal coherence via flip rate and lag-1 autocorrelation (Section~\ref{sec:fm4}).

\subsection{Prompting Strategies}
\label{sec:strategies}

Figure~\ref{fig:pipeline} illustrates the overall experimental pipeline.
We design and evaluate six prompting strategies of increasing complexity (Table~\ref{tab:strategy_summary}).

\begin{table}[t]
\centering
\small
\begin{tabular}{clcc}
\toprule
ID & Strategy & Theory & Retrieval \\
\midrule
S1 & Zero-shot & $\times$ & $\times$ \\
S2 & TG zero-shot & \checkmark & $\times$ \\
S3 & Few-shot (VLM) & $\times$ & VLM embedding \\
S4 & Few-shot (CLIP) & $\times$ & CLIP embedding \\
S5 & TG few-shot (VLM) & \checkmark & VLM embedding \\
S6 & TG few-shot (CLIP) & \checkmark & CLIP embedding \\
\bottomrule
\end{tabular}
\caption{Summary of the six prompting strategies.}
\label{tab:strategy_summary}
\end{table}

\subsubsection{S1: Zero-Shot Prediction}

The VLM receives all 16 frames from a 1-second gameplay window and a structured prompt that defines High and Low engagement in terms of observable visual cues (e.g., active combat and visible feedback vs.\ menus and empty environments), and asks the VLM to consider action intensity, HUD/UI elements, the presence of other entities, and signs of tension or reward (see Appendix~\ref{app:prompt_zeroshot} for the full prompt).
No demonstrations or theoretical guidance are provided.
This establishes the baseline for VLM's innate ability to judge engagement from short gameplay sequences.

\subsubsection{S2: Theory-Guided Zero-Shot}
\label{sec:theory_prompt}

We design a structured prompt that operationalizes the four theoretical frameworks from Section~\ref{sec:theories} into concrete visual indicators.
Each theory's core dimensions---e.g., challenge and feedback (Flow), control and immersion (GameFlow), competence and relatedness (SDT), sensation and fellowship (MDA)---are mapped to observable cues such as health bars, score displays, teammate visibility, and visual effects (see Appendix~\ref{app:prompt_theory} for the full prompt and complete dimension-to-cue mapping).
The VLM assesses each dimension, counts how many theories suggest High vs.\ Low, and outputs a label with a theory-grounded rationale.

\subsubsection{S3--S6: Retrieval-Augmented Strategies}
\label{sec:rag_strategies}

Strategies S3--S6 augment VLM predictions with retrieved demonstration examples \cite{yu2024visrag,mei2025multimodalrag}, varying along two orthogonal dimensions: \textit{embedding space} and \textit{theory guidance}.

\paragraph{Embedding space.}
We compare two embedding spaces for retrieval.
In S3 and S5, we extract embeddings from the VLM’s own last-layer hidden states, which capture the model’s internal representation of gameplay frames.
In S4 and S6, we use CLIP’s vision encoder \cite{radford2021clip}, which provides a task-agnostic visual-semantic embedding space.
Both embeddings are L2-normalized; retrieval is performed via cosine similarity, selecting the top-$k$ most similar cases as demonstrations.

\paragraph{Retrieval sources.}
For pointwise prediction, S3 and S5 use an in-game error-driven case base: we first run the model on 70\% of the windows from a target game, store incorrectly predicted cases in memory, and retrieve from this memory during inference on the remaining 30\%.
This design is intentional and targets the model’s specific weaknesses within a game; it should therefore be interpreted as a memory-based corrective adaptation setting rather than a strict held-out generalization benchmark.
For pairwise prediction, demonstrations are drawn from DOOM via difference-vector retrieval: we compute embedding differences between consecutive frame pairs and retrieve DOOM pairs with similar difference vectors, enabling cross-game transfer of engagement change patterns.
Pointwise and pairwise retrieval use different sources by design and should not be directly compared.

\paragraph{Theory-guided variants.}
In S5 and S6, retrieved demonstrations are incorporated into the structured theory prompt (Section~\ref{sec:theory_prompt}), combining concrete examples with theoretical scaffolding.

\section{Results: Do VLMs Understand Engagement?}
\label{sec:results}
We organize results around three research questions.
Tables~\ref{tab:pointwise_results} and~\ref{tab:pairwise_results} present the per-game accuracy breakdown and overall averages for each VLM across all prompting strategies.

\begin{table*}[t]
\centering
\small
\resizebox{\textwidth}{!}{%
\begin{tabular}{l c cccccc cccccc cc}
\toprule
 & & \multicolumn{6}{c}{\textbf{InternVL}} & \multicolumn{6}{c}{\textbf{Qwen}} & \multicolumn{2}{c}{\textbf{GPT-4o}} \\
\cmidrule(lr){3-8} \cmidrule(lr){9-14} \cmidrule(lr){15-16}
\textbf{Game} & M & S1 & S2 & S3 & S4 & S5 & S6 & S1 & S2 & S3 & S4 & S5 & S6 & S1 & S2 \\
\midrule
Borderlands 3  & 81.4 & 79.7 & 81.4 & \textbf{83.4} & 79.7 & 53.5 & 81.4 & 83.1 & 81.4 & 75.4 & 49.2 & \textbf{84.2} & 54.2 & \textbf{83.1} & 81.4 \\
CS:GO Office   & 66.1 & 61.0 & 42.4 & 58.5 & 55.9 & \textbf{71.8} & 33.9 & 57.6 & 44.1 & 71.1 & \textbf{86.4} & 70.8 & 78.0 & \textbf{52.5} & 50.8 \\
Blitz Brigade  & 64.4 & 66.1 & 64.4 & 68.7 & \textbf{69.5} & 62.7 & 64.4 & 62.7 & 64.4 & \textbf{78.9} & 66.1 & 78.2 & 66.1 & 62.7 & \textbf{64.4} \\
Corridor 7     & 61.0 & 49.2 & 37.3 & \textbf{83.5} & 64.4 & 72.9 & 39.0 & 66.1 & 39.0 & 72.5 & 55.9 & \textbf{73.9} & 52.5 & \textbf{57.6} & \textbf{57.6} \\
Battlefield 42 & 61.0 & 64.4 & 52.5 & 46.8 & 39.0 & \textbf{64.8} & 39.0 & 62.7 & 57.6 & \textbf{80.3} & 61.0 & 68.3 & 72.9 & \textbf{64.4} & 45.8 \\
Apex Legends   & 55.9 & 45.8 & 54.2 & \textbf{64.1} & 49.2 & 54.6 & 47.5 & 49.2 & 55.9 & 68.3 & \textbf{69.5} & 62.7 & 62.7 & \textbf{52.5} & \textbf{52.5} \\
CSGO19         & 59.3 & 71.2 & 59.3 & 76.8 & 50.8 & \textbf{82.0} & 64.4 & 66.1 & 54.2 & \textbf{84.2} & 55.9 & 83.8 & 71.2 & 55.9 & \textbf{74.6} \\
CSGO18         & 78.0 & 44.1 & 25.4 & \textbf{81.7} & 49.2 & 76.8 & 23.7 & 30.5 & 25.4 & 66.5 & 86.4 & 71.1 & \textbf{89.8} & 22.8 & \textbf{24.6} \\
CS 1.6         & 78.0 & 30.5 & 59.3 & \textbf{78.2} & 54.2 & 75.4 & 69.5 & 32.2 & 55.9 & 78.2 & \textbf{79.7} & 73.6 & \textbf{79.7} & \textbf{57.6} & \textbf{57.6} \\
\midrule
\textbf{Average} & 67.2 & 56.9 & 52.9 & 71.3 & 56.9 & 68.3 & 51.4 & 56.7 & 53.1 & 75.0 & 67.8 & 74.1 & 69.7 & 56.6 & 56.6 \\
\bottomrule
\end{tabular}%
}
\caption{Per-game pointwise engagement prediction accuracy (\%). M = per-game majority-class baseline. Boldface indicates the best result per game within each model. S3/S5 are evaluated on the $\approx$18-window held-out split, all other strategies on all 59 windows; 95\% CIs in Appendix~\ref{app:ci}.}
\label{tab:pointwise_results}
\end{table*}

\begin{table*}[t]
\centering
\small
\resizebox{\textwidth}{!}{%
\begin{tabular}{l c cccccc cccccc cc}
\toprule
 & & \multicolumn{6}{c}{\textbf{InternVL}} & \multicolumn{6}{c}{\textbf{Qwen}} & \multicolumn{2}{c}{\textbf{GPT-4o}} \\
\cmidrule(lr){3-8} \cmidrule(lr){9-14} \cmidrule(lr){15-16}
\textbf{Game} & M & S1 & S2 & S3 & S4 & S5 & S6 & S1 & S2 & S3 & S4 & S5 & S6 & S1 & S2 \\
\midrule
Borderlands 3  & 50.1 & \textbf{54.6} & 51.5 & 53.0 & 50.0 & 50.0 & 25.0 & 57.6 & 65.2 & 53.0 & 62.1 & \textbf{87.5} & \textbf{87.5} & 54.6 & \textbf{56.1} \\
CS:GO Office   & 58.7 & 42.7 & 49.3 & 42.7 & \textbf{52.0} & 23.1 & 30.8 & 37.3 & 45.3 & 46.7 & 42.7 & 69.2 & \textbf{76.9} & 41.3 & \textbf{45.3} \\
Blitz Brigade  & 57.6 & 56.1 & 47.0 & 65.1 & 57.6 & \textbf{71.4} & 64.3 & \textbf{68.2} & 62.1 & 51.5 & 51.5 & 57.1 & 42.9 & \textbf{65.2} & 60.6 \\
Corridor 7     & 64.9 & 51.4 & 40.5 & 51.4 & 67.6 & \textbf{70.0} & \textbf{70.0} & \textbf{73.0} & 51.4 & 35.1 & 64.9 & 50.0 & 50.0 & \textbf{59.5} & 56.8 \\
Battlefield 42 & 60.7 & \textbf{57.4} & 55.7 & 55.7 & 49.2 & 47.4 & 47.4 & 62.3 & 64.0 & 54.1 & \textbf{67.2} & 52.6 & 52.6 & 54.1 & \textbf{57.4} \\
Apex Legends   & 58.8 & 67.7 & 57.4 & 55.9 & 61.8 & \textbf{71.4} & 57.1 & 58.8 & \textbf{60.3} & 57.4 & 57.4 & 57.1 & 50.0 & 57.4 & \textbf{66.2} \\
CSGO19         & 52.2 & 43.5 & 52.2 & 50.0 & 41.3 & \textbf{70.0} & \textbf{70.0} & 45.7 & 45.7 & 43.5 & 41.3 & \textbf{50.0} & 40.0 & 43.5 & \textbf{52.2} \\
CSGO18         & 50.7 & 52.2 & 50.7 & 46.4 & \textbf{53.6} & 33.3 & 16.7 & 40.6 & 39.1 & 47.8 & 39.1 & \textbf{50.0} & \textbf{50.0} & 49.3 & \textbf{52.2} \\
CS 1.6         & 55.1 & 52.6 & 43.6 & 48.7 & 44.9 & \textbf{66.7} & \textbf{66.7} & 48.7 & 52.6 & 39.7 & 42.3 & \textbf{55.6} & 44.4 & 42.3 & \textbf{44.9} \\
\midrule
\textbf{Average} & 56.5 & 53.1 & 49.8 & 51.8 & 53.1 & 55.9 & 49.8 & 54.7 & 54.0 & 47.7 & 52.1 & 58.8 & 54.9 & 51.9 & 54.6 \\
\bottomrule
\end{tabular}%
}
\caption{Per-game pairwise engagement prediction accuracy (\%). M = per-game majority-class baseline. Boldface indicates the best result per game within each model. Per-game sample sizes range from 37 to 78 pairs; 95\% CIs in Appendix~\ref{app:ci}.}
\label{tab:pairwise_results}
\end{table*}

\subsection{Q1\&Q2: Can VLMs Predict Engagement, and Does Prompting Help?}
\label{sec:multi_game}

We evaluate all six prompting strategies across 9 games.
Pointwise and pairwise settings use different retrieval sources (Section~\ref{sec:rag_strategies}) and should be interpreted as complementary evaluations.

\paragraph{Overall accuracy is well below the majority-class baseline.}
Averaged across 9 games, zero-shot prediction achieves $\sim$57\% for all three models, well below the 67.2\% majority-class baseline.
GPT-4o achieves virtually identical zero-shot accuracy to the open-source models, suggesting that model scale alone does not resolve this difficulty.
Only a handful of game--strategy combinations exceed the per-game majority baseline, and performance varies dramatically: from 83.1\% on Borderlands~3 (vs.\ 81.4\% majority) to 30.5\% on CS~1.6 (vs.\ 78.0\% majority), echoing the game-dependence observed by \citet{melhart2025gamevibe}.

\paragraph{Theory-guided prompting provides mixed but often positive effects.}
Theory-guided zero-shot prompting (S2) alone does not consistently improve over plain zero-shot prompting (S1).
However, when combined with few-shot demonstrations (S5), theory-guided prompting often yields additional improvements over few-shot alone.
For example, S5 achieves 68.3\% accuracy for InternVL and 74.1\% for Qwen, suggesting that theory guidance is most effective when paired with concrete examples.
\paragraph{Embedding-based retrieval shows model-dependent effects.}
Retrieval-augmented prompting produces mixed results across models.
For InternVL, retrieval-based strategies provide limited improvement over few-shot prompting.
However, Qwen3-VL benefits substantially from CLIP-based retrieval, achieving 67.8\% with S4 and 69.7\% with S6, exceeding the majority-class baseline on average.
Post-hoc analysis confirms that scene-level visual similarity to the DOOM source domain strongly predicts retrieval-augmented accuracy (Appendix~\ref{app:retrieval_similarity}).

\paragraph{Statistical uncertainty.}
With only 59 windows per game, bootstrap 95\% confidence intervals span roughly $\pm$12pp, and most per-game differences between strategies have overlapping intervals (Appendix~\ref{app:ci}).
Aggregate patterns are more reliable: Qwen's retrieval-augmented average (S3: 75.0\% [71.4, 78.6]) is separated from its zero-shot average (S1: 56.7\% [46.0, 66.5]).

\subsection{Q2 (cont.): Qualitative Analysis of Theory-Aligned Rationales}

Adding theoretical structure does not consistently improve prediction: rationales tend to repeat surface observations (e.g., ``teammates visible $\Rightarrow$ relatedness $\Rightarrow$ High'') without genuine multi-dimensional reasoning (Section~\ref{sec:fm2}), suggesting information overload \cite{chochlakis2024strong}.
However, when theory prompts are combined with few-shot examples (S5, S6), improvements emerge---indicating that theoretical cues provide useful structure only when grounded by concrete examples.

\subsection{Q3: Are Predictions Consistent Across Games and VLMs?}

Both models consistently over-predict High engagement, but severity varies across games (Table~\ref{tab:error_dist}): on CSGO18, 90--97\% of errors are false positives, while CS:GO Office shows 60--74\%.
Cross-VLM agreement is substantial overall ($\kappa = 0.67$) but high agreement does not imply correctness---on CSGO18, models agree on 86.4\% of frames yet both achieve below-majority accuracy, converging on the same wrong answer.
This suggests both architectures rely on similar visual shortcuts.
We elaborate on specific failure modes in Section~\ref{sec:failure}.

\begin{table}[t]
\centering
\small
\begin{tabular}{llccc}
\toprule
\textbf{Game} & \textbf{Model} & \textbf{FN} & \textbf{FP} & \textbf{Total Err.} \\
\midrule
\multirow{2}{*}{CSGO18} & InternVL & 1 & 32 & 33 \\
                         & Qwen3-VL & 4 & 37 & 41 \\
\midrule
\multirow{2}{*}{CS:GO Office} & InternVL & 6 & 17 & 23 \\
                               & Qwen3-VL & 10 & 15 & 25 \\
\bottomrule
\end{tabular}
\caption{Zero-shot (S1) error distribution on the two CS:GO variants.}
\label{tab:error_dist}
\end{table}

\section{Analysis \& Failure Modes}
\label{sec:failure}

We conduct in-depth analysis on two CS:GO variants (CSGO18 and CS:GO Office), which exhibit contrasting error severities (Table~\ref{tab:error_dist}), identifying five systematic failure modes.

\subsection{Failure Mode 1: Visual Intensity Bias}
\label{sec:fm1}

VLMs systematically conflate visual intensity with engagement.
We quantify visual intensity as a composite of color saturation, edge density, brightness variance, and red channel intensity (Appendix~\ref{app:visual_intensity}).
On CSGO18, VLM predictions correlate significantly with this score ($r = 0.432$, $p < 0.001$), while human engagement labels show no correlation ($r = -0.193$, $p = 0.143$).
The VLM performs visual intensity classification rather than engagement prediction: it recognizes ``this frame looks intense'' but cannot reason that a 15--14 scoreboard implies competitive tension.
This bias is most pronounced when visual variation across frames is large; on CS:GO Office, neither VLM predictions nor human labels correlate with visual intensity---consistent with findings that VLMs perform best on action-heavy games \cite{melhart2025gamevibe} and systematically underutilize visual representations for reasoning \cite{fu2025hiddenplainsightvlms}.

\subsection{Failure Mode 2: Surface Feature Shortcuts}
\label{sec:fm2}

VLMs rely on surface-level visual shortcuts to predict engagement, bypassing genuine reasoning:

\paragraph{Relatedness bias.}
When teammates are visible or radio chatter indicators appear, the model automatically predicts High engagement.
The VLM equates the presence of social cues with high engagement, without considering whether team interaction is actually meaningful in the current context.

\paragraph{Sensation bias.}
Blood splatter, explosions, and weapon effects trigger High predictions regardless of gameplay context.
The model uses visual ``spectacle'' as a proxy for engagement.

These shortcuts map directly to theory dimensions (SDT's relatedness, MDA's sensation) but in a shallow, stimulus-response manner.
The theory-guided prompt amplifies these biases by providing vocabulary for rationalizing surface-level pattern matching, mirroring the sentiment bias in VLM-based emotion recognition \cite{bhattacharyya2025evaluatingvisionlanguagemodelsemotion}.

Annotating all 78 errors from S2 on CSGO18 and CS:GO Office reveals that Challenge Misassessment (38.5\%) and Relatedness Shortcuts (30.8\%) together account for over two-thirds of all errors (Appendix~\ref{app:error_taxonomy}).
A human evaluation of 30 sampled errors found that 40\% stem from clear reasoning flaws, while 60\% reflect defensible interpretations---highlighting the inherent difficulty of binary engagement classification.

\subsection{Failure Mode 3: Post-Match Context Blindness}
\label{sec:fm3}

VLMs cannot distinguish active gameplay from post-match screens.
Scoreboards showing close victories are high-engagement moments for human annotators, but VLMs classify them as ``static, no active gameplay'' and predict Low.
On CS:GO Office, 18 of 20 false negatives involve this misclassification---the model equates visual stillness with disengagement, missing that a scoreboard marks the culmination of a competitive experience.

\subsection{Failure Mode 4: Temporal Inconsistency}
\label{sec:fm4}

VLM predictions lack temporal coherence: the model may predict High for window $t$ and Low for window $t{+}1$ even when gameplay context is unchanged.
We measure this using flip rate (fraction of adjacent windows with different predictions) and lag-1 autocorrelation (Table~\ref{tab:temporal}).
On CSGO18, the VLM flip rate is 0.310---18$\times$ higher than the human label flip rate of 0.017---and its autocorrelation (0.275) is far below human labels (0.950).
This matches \citet{jung2025consistencyvideollm}'s finding that Video-LLMs achieve near chance-level consistency.

\begin{table}[t]
\centering
\small
\resizebox{\columnwidth}{!}{%
\begin{tabular}{llcc}
\toprule
\textbf{Metric} & \textbf{Dataset} & \textbf{VLM Pred.} & \textbf{Human Labels} \\
\midrule
\multirow{2}{*}{Flip Rate $\downarrow$} & CSGO18 & 0.310 & 0.017 \\
& CS:GO Office & 0.345 & 0.190 \\
\midrule
\multirow{2}{*}{Autocorr. (lag-1) $\uparrow$} & CSGO18 & 0.275 & 0.950 \\
& CS:GO Office & 0.168 & 0.576 \\
\bottomrule
\end{tabular}%
}
\caption{Temporal coherence metrics, averaged over strategies S1--S6 and the two open-source VLMs. VLM predictions are dramatically less coherent than human labels across consecutive 1-second windows.}
\label{tab:temporal}
\end{table}

\subsection{Failure Mode 5: Confident but Wrong Reasoning}
\label{sec:fm5}

VLMs generate detailed, fluent, and seemingly well-reasoned rationales that are nonetheless incorrect, performing post-hoc rationalization of surface-level cues rather than genuine engagement reasoning \cite{li2026hallucination}.
Strikingly, we find an inverse confidence--accuracy relationship: high-confidence predictions achieve only 38.5\% accuracy, while low-confidence predictions achieve 64.3\% (Appendix~\ref{app:error_viz}), confirming that confident VLM reasoning is systematically misleading.
Four orderings of few-shot examples yield nearly identical prediction distributions (Appendix~\ref{app:order}), confirming an intrinsic model bias rather than a positional artifact.

\subsection{Analysis: Demonstration Label Balance}
\label{sec:label_balance}

Beyond failure modes, we find that few-shot predictions are highly sensitive to the label distribution of retrieved demonstrations.
In our error-driven retrieval setup (Section~\ref{sec:rag_strategies}), several games' case bases contain no positive examples, causing systematic over-prediction of the negative class.
Adding even a small number of positive cases dramatically improves accuracy (e.g., +15.5pp on Blitz Brigade with just 17 examples; Appendix~\ref{app:label_balance}).
This implies that retrieval-augmented VLM systems for subjective tasks must ensure balanced demonstration selection.

\section{Discussion}
\label{sec:discussion}

\subsection{The Perception--Understanding Gap}

Our failure modes collectively point to a single underlying issue: VLMs treat engagement prediction as visual classification rather than psychological inference, stopping at the first step of the required multi-step reasoning chain \cite{fu2025hiddenplainsightvlms}.
Games are uniquely diagnostic because they decouple visual salience from the target construct.
Our results suggest a capability hierarchy---perception $>$ description $>$ interpretation $>$ inference---where current VLMs succeed at the first two levels but fail at the latter two.

\subsection{The Role of Prompting Complexity}
\label{sec:prompting_complexity}

The interaction between prompting strategy and performance reveals three key findings.
First, few-shot demonstrations provide large, reliable pointwise gains, though these do not extend to pairwise prediction---likely because cross-game DOOM demonstrations are less informative than in-game memory.
Second, theory guidance alone hurts performance, likely through shortcut amplification: explicitly naming features like ``teammates visible'' provides vocabulary for rationalizing surface-level pattern matching rather than enabling genuine reasoning (Section~\ref{sec:fm2}).
Third, concrete examples drive improvements more than theoretical scaffolding: when combined with few-shot demonstrations, theory guidance provides at best modest additional benefit.
Cross-game retrieval effectiveness further depends on scene-level visual similarity rather than game-level semantic similarity (Appendix~\ref{app:retrieval_similarity}), and few-shot predictions are sensitive to demonstration label balance (Appendix~\ref{app:label_balance}).

\subsection{Implications}

Our findings argue for VLM benchmarks that test affective understanding beyond perception, especially as VLMs are increasingly deployed as automated judges and rankers of subjective content \cite{wang2026streetdesignai,du2026multimodalgeo,xing2026rankers}; game engagement offers a particularly diagnostic testbed.
For game analytics, VLMs are not yet suitable without task-specific fine-tuning via LoRA \cite{hu2022lora} or game-specific retrieval \cite{lewis2020rag}.
For prompt engineering, our results caution that more structured prompts may amplify biases rather than compensate for them when the model lacks genuine understanding.

\paragraph{Recommendations.}
Future work should (1)~decouple perception from inference in evaluation; (2)~prioritize concrete examples over elaborate prompts for subjective tasks; (3)~incorporate temporal context via memory mechanisms \cite{nian2025jaildam}; and (4)~calibrate VLM confidence (Section~\ref{sec:fm5}).
We outline concrete architectural directions in Appendix~\ref{app:solutions}.

\section{Conclusion}
\label{sec:conclusion}

We presented the first systematic evaluation of VLMs on human engagement prediction in gameplay videos, testing six prompting strategies across 9 games with three VLMs.
Our results show that raw zero-shot predictions are unreliable ($\sim$57\%, below the majority-class baseline), but few-shot demonstrations yield substantial pointwise gains (up to 75.0\% for Qwen), while theory-guided prompting alone can amplify surface-level shortcuts rather than improve reasoning.
Our failure-mode analysis reveals five systematic limitations: VLMs conflate visual intensity with engagement, rely on surface-feature shortcuts, cannot interpret post-match contexts, lack temporal coherence, and generate confident but misleading rationales---collectively exposing a fundamental perception--understanding gap that standard benchmarks would not detect.
Games, where visual salience is decoupled from the target construct, provide a uniquely diagnostic testbed for this gap.
Notably, pairwise prediction remains near chance level across all strategies, suggesting that using VLMs as automated preference annotators for subjective tasks may be unreliable without substantial task-specific adaptation.
We outline concrete architectural directions for bridging this gap in Appendix~\ref{app:solutions}.

\section*{Limitations}

Our evaluation is limited by the GVFS dataset's scale (61 seconds per game, 59 windows each) and genre diversity (FPS only).
Engagement labels come from a pool of 20 trained observers (5 per clip) whose demographic and gaming-background profiles are unavailable, so annotator characteristics (e.g., sensitivity to visual spectacle) cannot be disentangled from model biases.
Binarization of continuous engagement scores may lose meaningful variation; finer-grained approaches could reveal different capabilities.
We evaluate two open-source VLMs with all six strategies but only test GPT-4o on S1--S2 due to cost constraints; broader model coverage could provide further insights.

\section*{Ethics Statement}

This study uses the publicly available GameVibe Few-Shot dataset, which contains gameplay video clips annotated for engagement.
No personally identifiable information is involved; all annotations are aggregated and anonymized.
Our work evaluates existing pretrained VLMs without fine-tuning and does not introduce new models or training data.
The engagement labels reflect subjective judgments from a small annotator pool, and we acknowledge that binarization may oversimplify the nuanced nature of engagement.
Because the annotators were gaming-familiar observers, sensitivity to visual spectacle may differ for other populations (e.g., minors or viewers less habituated to FPS content), and our conclusions should not be extrapolated to such groups.

\section*{Acknowledgments}

The authors acknowledge Research Computing at Arizona State University \cite{jennewein2023sol} for providing GPU computing resources that have contributed to the research results reported within this paper.

\bibliography{custom}

\appendix

\section{Error Taxonomy and Human Evaluation}
\label{app:error_taxonomy}

Three annotators independently classified all 78 prediction errors from S2 (InternVL3.5-8B-Instruct) on CSGO18 and CS:GO Office using a two-stage procedure: (1)~keyword-based candidate identification from VLM rationales, and (2)~manual review and correction. Inter-annotator agreement was moderate (Fleiss' $\kappa = 0.512$); disagreements were resolved by majority vote. Table~\ref{tab:error_breakdown} shows the resulting taxonomy.

\begin{table}[ht]
\centering
\small
\resizebox{\columnwidth}{!}{%
\begin{tabular}{lcccc}
\toprule
\textbf{Failure Type} & \textbf{CSGO18} & \textbf{CS:GO Office} & \textbf{Total} & \textbf{\%} \\
\midrule
A: Static Scene Misclass. & 3 & 4 & 7 & 9.0 \\
B: Relatedness Shortcut & 19 & 5 & 24 & 30.8 \\
C: Sensation Shortcut & 0 & 10 & 10 & 12.8 \\
D: Challenge Misassess. & 16 & 14 & 30 & 38.5 \\
E: Other/Ambiguous & 6 & 1 & 7 & 9.0 \\
\midrule
Total Errors & 44 & 34 & 78 & \\
\bottomrule
\end{tabular}%
}
\caption{Distribution of failure types. Challenge Misassessment and Relatedness Shortcuts account for over two-thirds of all errors.}
\label{tab:error_breakdown}
\end{table}

\paragraph{Human evaluation of error severity.}
Three annotators judged a stratified sample of 30 errors as Model Wrong (MW), Ground Truth Ambiguous (GA), or Both Plausible (BP). Inter-annotator agreement: Fleiss' $\kappa = 0.679$.
Of 30 cases, 12 (40\%) were MW and 18 (60\%) were BP, with no GA cases.

\section{Visual Intensity Score}
\label{app:visual_intensity}

We quantify visual intensity for each frame as a composite score:
\begin{equation}
I(f) = \frac{1}{4}\left(\hat{S}(f) + \hat{E}(f) + \hat{V}(f) + \hat{R}(f)\right)
\end{equation}
where $\hat{S}$, $\hat{E}$, $\hat{V}$, and $\hat{R}$ are the min-max normalized values of color saturation, Canny edge density, brightness variance, and red channel intensity, respectively.
On CSGO18, the mean visual intensity for frames predicted as High (0.220) is significantly higher than for Low (0.177), whereas ground-truth High (0.189) and Low (0.211) frames show no such separation (Spearman $\rho = 0.490$, $p < 0.001$ for VLM predictions; $\rho = -0.193$, $p = 0.143$ for human labels).

\section{Retrieval Similarity Analysis}
\label{app:retrieval_similarity}

The effectiveness of retrieval-augmented strategies depends on visual similarity between the source (DOOM) and target game.
We compare two CS:GO variants (Table~\ref{tab:retrieval_similarity}): CS:GO Office (indoor, DOOM-like) achieves 63.9\% retrieval-augmented accuracy with mean CLIP similarity of 0.657, while CSGO18 (outdoor) achieves only 37.7\% with similarity 0.628.
This 4.6\% gap is highly significant ($t(120) = 5.09$, $p < 10^{-5}$; Cohen's $d = 0.93$) and corresponds to a 26.2pp accuracy difference ($\chi^2 = 7.38$, $p = 0.007$).
Per-frame correlation between similarity and correctness is weak (CS:GO Office: $r = 0.14$; CSGO18: $r = 0.01$), suggesting the effect operates at the game level.

\begin{table}[ht]
\centering
\small
\resizebox{\columnwidth}{!}{%
\begin{tabular}{lcccc}
\toprule
\textbf{Target Game} & \textbf{Avg Sim.} & \textbf{SD} & \textbf{Acc.\ (\%)} & \textbf{$r$} \\
\midrule
CS:GO Office & 0.657 & 0.030 & 63.9 & 0.14 \\
CSGO18 & 0.628 & 0.033 & 37.7 & 0.01 \\
\midrule
\multicolumn{2}{l}{Similarity: $t(120) = 5.09$} & \multicolumn{3}{l}{$p < 10^{-5}$, Cohen's $d = 0.93$} \\
\multicolumn{2}{l}{Accuracy: $\chi^2 = 7.38$} & \multicolumn{3}{l}{$p = 0.007$} \\
\bottomrule
\end{tabular}%
}
\caption{CLIP retrieval similarity and accuracy for two CS:GO variants.}
\label{tab:retrieval_similarity}
\end{table}

\begin{figure}[ht]
\centering
\includegraphics[width=\columnwidth]{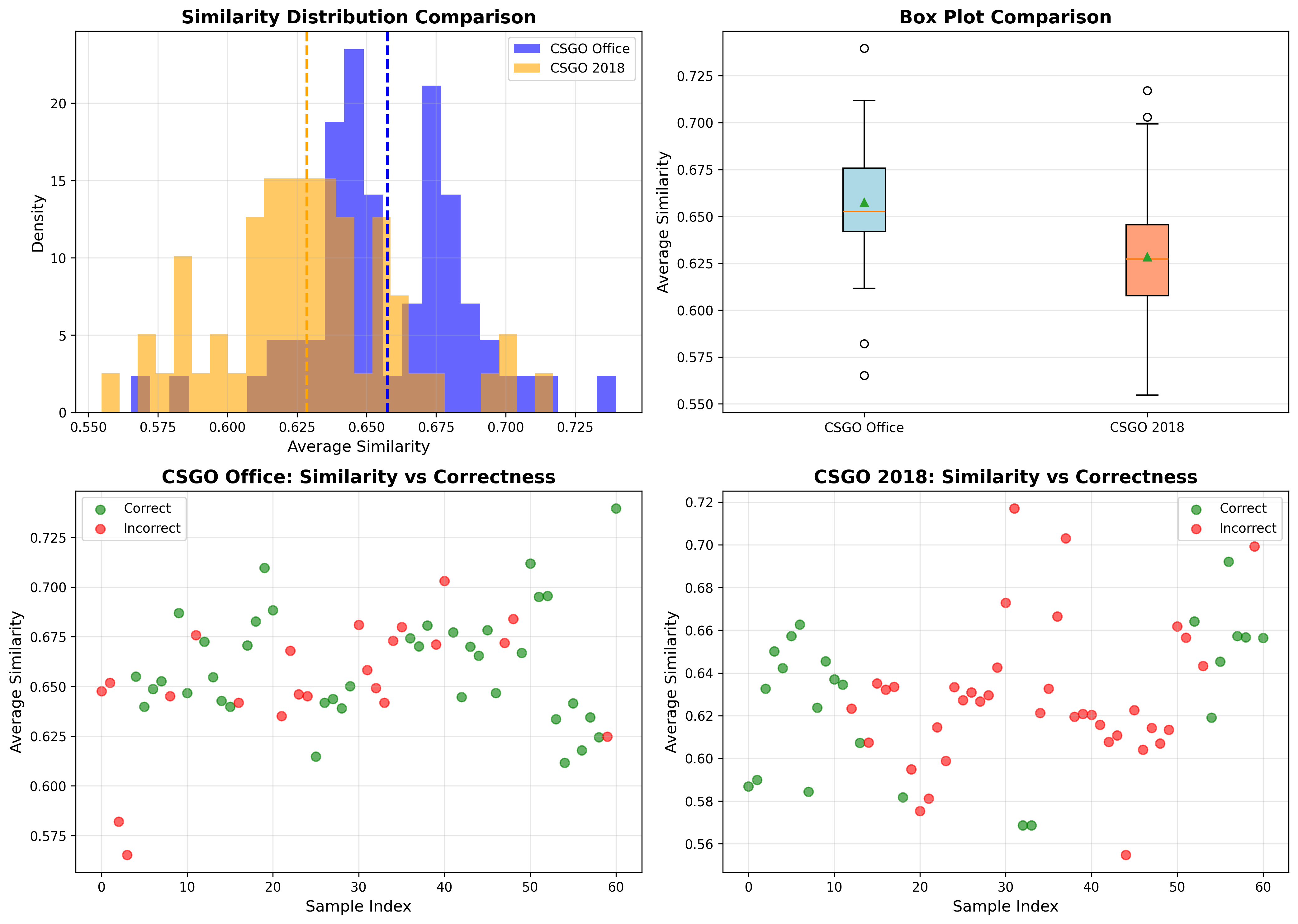}
\caption{CLIP retrieval similarity analysis for two CS:GO variants. \textbf{Top}: Similarity distributions and box plots. \textbf{Bottom}: Per-frame similarity vs.\ prediction correctness.}
\label{fig:retrieval_comparison}
\end{figure}

\section{Demonstration Label Balance}
\label{app:label_balance}

Few-shot VLM predictions are highly sensitive to the label distribution of demonstration examples.
In our few-shot setup, we first run the model on 70\% of the windows from a game and store incorrectly predicted cases in an error-driven memory.
During inference on the remaining 30\%, the model retrieves examples from this memory as few-shot demonstrations.
Because the memory is constructed only from the earlier split, no evaluation samples are included in the demonstration set.

\begin{figure}[ht]
\centering
\includegraphics[width=\columnwidth]{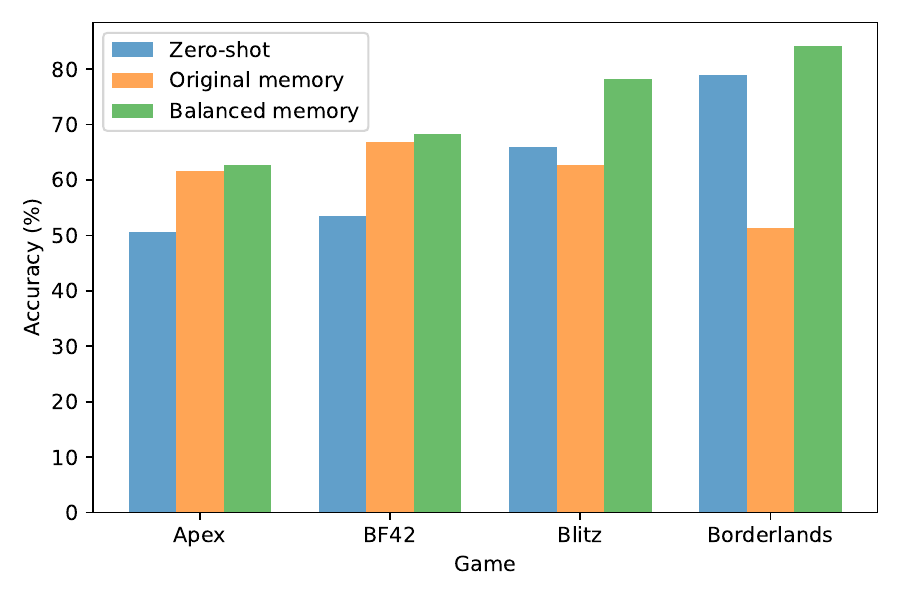}
\caption{Effect of demonstration label distribution on few-shot prediction accuracy. When the original memory contains no positive examples, the model over-predicts the negative class. Introducing positive examples into the demonstration memory consistently improves accuracy across multiple games.}
\label{fig:label_balance}
\end{figure}

For several games, the original memory contained no positive examples (label = 1), resulting in a highly imbalanced demonstration set where the model observes only negative examples during in-context learning, causing it to over-predict the negative class.
To investigate this effect, we introduced a small number of positive examples into the memory.
Adding only 17 positive cases---corresponding to the average number of positive examples available in other games' memories---consistently improved accuracy (e.g., Blitz Brigade: 62.7\% $\to$ 78.2\%; Borderlands~3: 51.4\% $\to$ 66.5\%), and increasing to 50 positive examples raised accuracy to 84.2\% (Figure~\ref{fig:label_balance}).

\section{Bootstrap Confidence Intervals and Evaluation-Subset Effects}
\label{app:ci}

Given the small per-game evaluation sets, we quantify the uncertainty of all accuracy estimates in Tables~\ref{tab:pointwise_results} and~\ref{tab:pairwise_results} with bootstrap 95\% confidence intervals (10{,}000 resamples).
For per-game entries we resample the per-window (or per-pair) correctness outcomes; for the cross-game averages we resample the nine per-game accuracies, so those intervals reflect variation across games.
Pointwise strategies are evaluated on $n{=}59$ windows per game; pairwise sample sizes vary by game because near-stable pairs are excluded (Borderlands~3: 66, CS:GO Office: 75, Blitz Brigade: 66, Corridor~7: 37, Battlefield~42: 61, Apex Legends: 68, CSGO19: 46, CSGO18: 69, CS~1.6: 78 pairs).
Because S3/S5 (pointwise) and S5/S6 (pairwise) are evaluated on smaller strategy-specific subsets, their per-game intervals are not directly comparable to the $n{=}59$ columns, and we report game-level (Average-column) intervals only.
Tables~\ref{tab:ci_pointwise} and~\ref{tab:ci_pairwise} report the intervals.

Two observations follow.
First, per-game intervals at $n{=}59$ span roughly $\pm$12pp, so most per-game differences between prompting strategies fall within overlapping intervals; only differences exceeding roughly 23pp separate reliably at this scale.
Second, aggregate patterns are more robust: Qwen's retrieval-augmented average remains outside its zero-shot interval (S3: 75.0 [71.4, 78.6] vs.\ S1: 56.7 [46.0, 66.5]), whereas in the pairwise setting every strategy's interval overlaps the majority-class baseline, reinforcing the conclusion that pairwise engagement-change prediction remains at chance level.

Because S3/S5 evaluate on held-out subsets of $\approx$18 windows while S1/S2/S4/S6 evaluate on all 59 windows, point comparisons across these columns mix evaluation sets.
Table~\ref{tab:subset_effect} quantifies this subset effect for the zero-shot strategies: for each game we draw 10{,}000 random 18-window subsets of the 59 evaluation windows and report the mean and central 95\% range of S1/S2 accuracy at the S3/S5 evaluation size.
Subset means coincide with the full-set accuracies, so the full-set S1/S2 numbers are unbiased points of comparison; the wide subset ranges ($\pm$18pp) nevertheless caution against reading small cross-column differences as meaningful.

\begin{table*}[t]
\centering
\small
\resizebox{\textwidth}{!}{%
\begin{tabular}{ll ccccccccc c}
\toprule
 & & \textbf{Bord.~3} & \textbf{Office} & \textbf{Blitz} & \textbf{Corr.~7} & \textbf{Battl.~42} & \textbf{Apex} & \textbf{CSGO19} & \textbf{CSGO18} & \textbf{CS 1.6} & \textbf{Average} \\
\midrule
\multirow{6}{*}{InternVL} & S1 & [69.5, 89.8] & [49.2, 72.9] & [54.2, 78.0] & [37.3, 62.7] & [52.5, 76.3] & [32.2, 59.3] & [59.3, 83.1] & [32.2, 57.6] & [18.6, 42.4] & 56.9 [47.3, 66.3] \\
 & S2 & [71.2, 91.5] & [30.5, 55.9] & [52.5, 76.3] & [25.4, 49.2] & [40.7, 66.1] & [40.7, 66.1] & [47.5, 71.2] & [15.3, 37.3] & [45.8, 71.2] & 52.9 [42.9, 62.9] \\
 & S3 & -- & -- & -- & -- & -- & -- & -- & -- & -- & 71.3 [63.0, 78.6] \\
 & S4 & [69.5, 89.8] & [42.4, 67.8] & [57.6, 81.4] & [52.5, 76.3] & [27.1, 50.8] & [37.3, 62.7] & [39.0, 62.7] & [37.3, 62.7] & [40.7, 66.1] & 56.9 [49.7, 64.8] \\
 & S5 & -- & -- & -- & -- & -- & -- & -- & -- & -- & 68.3 [62.0, 74.3] \\
 & S6 & [71.2, 91.5] & [22.0, 45.8] & [52.5, 76.3] & [27.1, 52.5] & [27.1, 52.5] & [33.9, 61.0] & [52.5, 76.3] & [13.6, 35.6] & [57.6, 81.4] & 51.4 [39.9, 63.3] \\
\midrule
\multirow{6}{*}{Qwen} & S1 & [72.9, 91.5] & [45.8, 69.5] & [50.8, 74.6] & [54.2, 78.0] & [50.8, 74.6] & [35.6, 62.7] & [54.2, 78.0] & [18.6, 42.4] & [20.3, 44.1] & 56.7 [46.0, 66.5] \\
 & S2 & [71.2, 91.5] & [32.2, 57.6] & [52.5, 76.3] & [27.1, 50.8] & [45.8, 69.5] & [44.1, 67.8] & [40.7, 66.1] & [15.3, 37.3] & [42.4, 69.5] & 53.1 [43.5, 63.1] \\
 & S3 & -- & -- & -- & -- & -- & -- & -- & -- & -- & 75.0 [71.4, 78.6] \\
 & S4 & [37.3, 62.7] & [76.3, 94.9] & [54.2, 78.0] & [42.4, 67.8] & [49.2, 72.9] & [57.6, 81.4] & [42.4, 67.8] & [78.0, 94.9] & [69.5, 89.8] & 67.8 [59.7, 76.4] \\
 & S5 & -- & -- & -- & -- & -- & -- & -- & -- & -- & 74.1 [69.8, 78.5] \\
 & S6 & [40.7, 66.1] & [67.8, 88.1] & [54.2, 78.0] & [39.0, 64.4] & [61.0, 83.1] & [50.8, 74.6] & [59.3, 83.1] & [81.4, 96.6] & [69.5, 89.8] & 69.7 [62.1, 77.0] \\
\midrule
\multirow{2}{*}{GPT-4o} & S1 & [72.9, 91.5] & [39.0, 66.1] & [50.8, 74.6] & [45.8, 69.5] & [52.5, 76.3] & [39.0, 64.4] & [42.4, 67.8] & [11.9, 33.9] & [45.8, 69.5] & 56.6 [46.2, 65.7] \\
 & S2 & [71.2, 91.5] & [39.0, 62.7] & [52.5, 76.3] & [44.1, 69.5] & [33.9, 59.3] & [39.0, 64.4] & [62.7, 84.7] & [13.6, 35.6] & [44.1, 69.5] & 56.6 [46.3, 66.7] \\
\bottomrule
\end{tabular}%
}
\caption{Bootstrap 95\% confidence intervals for the pointwise accuracies of Table~\ref{tab:pointwise_results}. Per-game entries resample the 59 per-window correctness outcomes; the Average column reports the cross-game mean with a game-level bootstrap interval (resampling the nine per-game accuracies). For S3/S5, which are evaluated on smaller held-out subsets ($\approx$30\% of windows), only the game-level interval is reported (see text).}
\label{tab:ci_pointwise}
\end{table*}

\begin{table*}[t]
\centering
\small
\resizebox{\textwidth}{!}{%
\begin{tabular}{ll ccccccccc c}
\toprule
 & & \textbf{Bord.~3} & \textbf{Office} & \textbf{Blitz} & \textbf{Corr.~7} & \textbf{Battl.~42} & \textbf{Apex} & \textbf{CSGO19} & \textbf{CSGO18} & \textbf{CS 1.6} & \textbf{Average} \\
\midrule
\multirow{6}{*}{InternVL} & S1 & [42.4, 66.7] & [32.0, 54.7] & [43.9, 68.2] & [35.1, 67.6] & [44.3, 68.9] & [55.9, 79.4] & [30.4, 58.7] & [40.6, 63.8] & [41.0, 64.1] & 53.1 [48.6, 57.8] \\
 & S2 & [39.4, 63.6] & [37.3, 60.0] & [34.8, 59.1] & [24.3, 56.8] & [42.6, 68.9] & [45.6, 69.1] & [37.0, 67.4] & [39.1, 62.3] & [32.1, 55.1] & 49.8 [46.4, 53.0] \\
 & S3 & [40.9, 65.2] & [32.0, 54.7] & [53.0, 75.8] & [35.1, 67.6] & [42.6, 67.2] & [44.1, 67.6] & [34.8, 65.2] & [34.8, 58.0] & [37.2, 60.3] & 51.8 [48.3, 56.3] \\
 & S4 & [37.9, 62.1] & [40.0, 62.7] & [45.5, 69.7] & [51.4, 81.1] & [37.7, 62.3] & [50.0, 73.5] & [28.3, 56.5] & [42.0, 65.2] & [34.6, 56.4] & 53.1 [48.1, 58.3] \\
 & S5 & -- & -- & -- & -- & -- & -- & -- & -- & -- & 55.9 [44.1, 66.2] \\
 & S6 & -- & -- & -- & -- & -- & -- & -- & -- & -- & 49.8 [36.6, 61.8] \\
\midrule
\multirow{6}{*}{Qwen} & S1 & [45.5, 69.7] & [26.7, 48.0] & [56.1, 78.8] & [56.8, 86.5] & [49.2, 73.8] & [47.1, 70.6] & [30.4, 60.9] & [29.0, 52.2] & [37.2, 60.3] & 54.7 [47.2, 62.2] \\
 & S2 & [53.0, 75.8] & [34.7, 56.0] & [50.0, 74.2] & [35.1, 67.6] & [52.5, 75.4] & [48.5, 72.1] & [30.4, 60.9] & [27.5, 50.7] & [41.0, 64.1] & 54.0 [48.1, 59.7] \\
 & S3 & [40.9, 65.2] & [36.0, 57.3] & [39.4, 63.6] & [18.9, 51.4] & [41.0, 65.6] & [45.6, 69.1] & [30.4, 58.7] & [36.2, 59.4] & [29.5, 51.3] & 47.7 [43.1, 52.0] \\
 & S4 & [50.0, 72.7] & [30.7, 53.3] & [39.4, 63.6] & [48.6, 78.4] & [55.7, 78.7] & [45.6, 69.1] & [28.3, 56.5] & [27.5, 50.7] & [32.1, 53.8] & 52.1 [45.4, 58.9] \\
 & S5 & -- & -- & -- & -- & -- & -- & -- & -- & -- & 58.8 [52.5, 67.4] \\
 & S6 & -- & -- & -- & -- & -- & -- & -- & -- & -- & 54.9 [46.1, 65.6] \\
\midrule
\multirow{2}{*}{GPT-4o} & S1 & [42.4, 66.7] & [30.7, 52.0] & [53.0, 75.8] & [43.2, 75.7] & [42.6, 65.6] & [45.6, 69.1] & [30.4, 58.7] & [37.7, 60.9] & [32.1, 52.6] & 51.9 [46.7, 57.1] \\
 & S2 & [43.9, 68.2] & [34.7, 56.0] & [48.5, 72.7] & [40.5, 73.0] & [44.3, 68.9] & [54.4, 76.5] & [37.0, 67.4] & [40.6, 63.8] & [33.3, 56.4] & 54.6 [50.5, 58.8] \\
\bottomrule
\end{tabular}%
}
\caption{Bootstrap 95\% confidence intervals for the pairwise accuracies of Table~\ref{tab:pairwise_results}. Per-game entries resample the per-pair correctness outcomes with the per-game pair counts given in the text; the Average column reports the cross-game mean with a game-level bootstrap interval. For S5/S6, which are evaluated on smaller strategy-specific subsets of pairs, only the game-level interval is reported.}
\label{tab:ci_pairwise}
\end{table*}

\begin{table*}[t]
\centering
\small
\resizebox{\textwidth}{!}{%
\begin{tabular}{l cccc cccc}
\toprule
 & \multicolumn{4}{c}{\textbf{InternVL}} & \multicolumn{4}{c}{\textbf{Qwen}} \\
\cmidrule(lr){2-5} \cmidrule(lr){6-9}
\textbf{Game} & S1 (18-win.\ subsets) & S2 (18-win.\ subsets) & S3 & S5 & S1 (18-win.\ subsets) & S2 (18-win.\ subsets) & S3 & S5 \\
\midrule
Borderlands 3 & 79.6 [61.1, 94.4] & 81.3 [66.7, 94.4] & 83.4 & 53.5 & 83.0 [66.7, 94.4] & 81.4 [66.7, 94.4] & 75.4 & 84.2 \\
CS:GO Office & 61.2 [44.4, 77.8] & 42.3 [22.2, 61.1] & 58.5 & 71.8 & 57.7 [38.9, 77.8] & 44.0 [27.8, 61.1] & 71.1 & 70.8 \\
Blitz Brigade & 66.1 [50.0, 83.3] & 64.5 [44.4, 83.3] & 68.7 & 62.7 & 62.7 [44.4, 83.3] & 64.5 [44.4, 83.3] & 78.9 & 78.2 \\
Corridor 7 & 49.4 [27.8, 66.7] & 37.3 [16.7, 55.6] & 83.5 & 72.9 & 66.1 [50.0, 83.3] & 38.8 [22.2, 55.6] & 72.5 & 73.9 \\
Battlefield 42 & 64.4 [44.4, 83.3] & 52.6 [33.3, 72.2] & 46.8 & 64.8 & 62.7 [44.4, 83.3] & 57.7 [38.9, 77.8] & 80.3 & 68.3 \\
Apex Legends & 45.8 [27.8, 66.7] & 54.1 [33.3, 72.2] & 64.1 & 54.6 & 49.1 [27.8, 66.7] & 56.0 [38.9, 77.8] & 68.3 & 62.7 \\
CSGO19 & 71.2 [55.6, 88.9] & 59.5 [38.9, 77.8] & 76.8 & 82.0 & 66.0 [50.0, 83.3] & 54.4 [33.3, 72.2] & 84.2 & 83.8 \\
CSGO18 & 44.1 [27.8, 61.1] & 25.5 [11.1, 44.4] & 81.7 & 76.8 & 30.5 [11.1, 50.0] & 25.4 [11.1, 44.4] & 66.5 & 71.1 \\
CS 1.6 & 30.5 [11.1, 50.0] & 59.3 [38.9, 77.8] & 78.2 & 75.4 & 32.2 [16.7, 50.0] & 55.9 [38.9, 72.2] & 78.2 & 73.6 \\
\bottomrule
\end{tabular}%
}
\caption{Evaluation-subset effect for the zero-shot strategies. For each game we draw 10{,}000 random 18-window subsets of the 59 evaluation windows and report the mean and central 95\% range of S1/S2 accuracy on subsets of the size used by the S3/S5 held-out evaluation; the S3/S5 columns repeat the accuracies from Table~\ref{tab:pointwise_results} for reference. Subset means coincide with the full-set accuracies, and the wide ranges indicate that point comparisons across evaluation subsets of this size carry substantial sampling variability.}
\label{tab:subset_effect}
\end{table*}

\section{Toward Practical Engagement Prediction Systems}
\label{app:solutions}

Our evaluation identifies clear limitations of raw VLMs for engagement prediction, but also reveals specific capabilities that can be leveraged through architectural augmentation.
Below we outline five concrete directions, each grounded in specific failure modes identified in Section~\ref{sec:failure}.

\paragraph{1. Hybrid pipeline: VLM features + supervised classifier.}
Use the VLM as a feature extractor---prompting it to output structured descriptions of observable game state (score, health, number of visible players, action level, UI state)---and then train a lightweight supervised classifier on the extracted features using the engagement annotations.

\paragraph{2. Temporal smoothing of per-window predictions.}
A post-processing temporal model---such as a hidden Markov model, exponential moving average, or a lightweight LSTM operating on the sequence of per-window VLM predictions and extracted features---could enforce temporal smoothness.
Given the strong autocorrelation in human engagement labels ($r_1 = 0.576$--$0.950$), even simple smoothing could substantially reduce flip rate.

\paragraph{3. Decomposed reasoning via chain-of-thought.}
Decompose the task into a pipeline: (i)~identify game state (active gameplay, menu, scoreboard, buy phase); (ii)~assess challenge level; (iii)~evaluate feedback; (iv)~predict engagement conditioned on (i)--(iii).

\paragraph{4. Game-specific calibration.}
Techniques such as Platt scaling \cite{platt1999probabilistic} or temperature scaling \cite{guo2017calibration}, applied to VLM prediction logits using as few as 20--30 labeled windows per game, could correct for game-specific prediction biases.

\paragraph{5. Structured output with verification.}
Prompt the VLM to extract a structured game-state representation (game phase, score differential, health, visible players, action level), which can then be processed by deterministic engagement logic.

\section{Zero-Shot Prompt}
\label{app:prompt_zeroshot}

Below is the complete prompt used in S1 (Section~\ref{sec:strategies}).

\small
\begin{lstlisting}
Task: Predict whether these game video frames
indicate High or Low player engagement.

You are given 16 frames from a 1-second gameplay
window. Analyze the sequence as a whole.

Engagement definition:
- High engagement: The player is actively involved
  in intense, meaningful, or challenging gameplay.
  Visual signs include active combat, time pressure,
  dynamic action, visible feedback from the game,
  interaction with other players, or tense
  situations.
- Low engagement: The player is in a passive, idle,
  or low-intensity state. Visual signs include
  menus/pause screens, empty environments, lack of
  action or feedback, repetitive/uneventful scenes,
  or the player appearing disengaged from core
  gameplay.

Look at the frames carefully and consider:
- What is happening in the scene? (action intensity,
  events)
- What HUD/UI elements are visible? (health, score,
  timer, kill feed, objectives)
- Are there other players or entities? (teammates,
  enemies, NPCs)
- Does the scene suggest tension, challenge, or
  reward?

OUTPUT (strict format):
Label: [High/Low]
Rationale: [One sentence explaining the key visual
evidence for your prediction]
\end{lstlisting}
\normalsize

\section{Full Theory-Guided Prompt}
\label{app:prompt_theory}

Below is the complete prompt used in S2, S5, and S6 (Section~\ref{sec:theory_prompt}).

\small
\begin{lstlisting}
Task: Predict whether these game video frames
indicate High or Low player engagement, using game
engagement theories as an analytical framework.

You are given 16 frames from a 1-second gameplay
window. Analyze the sequence as a whole.

ENGAGEMENT DEFINITION:
- High engagement = the player is deeply invested in
  gameplay -- facing meaningful challenge, receiving
  feedback, experiencing immersion, or interacting
  with others.
- Low engagement = the player is disengaged -- idle,
  in menus, facing no challenge, receiving no
  feedback, or in an uneventful situation.

THEORETICAL FRAMEWORK FOR VISUAL ANALYSIS:

1. Flow Theory (challenge-skill balance, goals,
   feedback):
   Look for signs of:
   - Active challenge: health not full/not empty
     (mid-range), score is close, enemies present,
     time pressure visible
   - Clear goals: objective markers, mission
     indicators, timers, waypoints on screen
   - Immediate feedback: damage numbers, hit
     markers, kill notifications, score changes,
     reward popups
   -> Low engagement signs: no visible objectives,
      no feedback elements, health full with no
      threats

2. GameFlow (control, immersion, social presence):
   Look for signs of:
   - Player control: crosshair/aim on target,
     character in motion, weapon/tool actively in use
   - Immersion: full gameplay view (no pause menus,
     no loading screens, no settings), rich
     environment detail
   - Social presence: other players visible, team
     indicators, chat activity, cooperative/
     competitive interaction
   -> Low engagement signs: pause/menu screens,
      empty environments, no other players or
      entities

3. Self-Determination Theory (competence, autonomy,
   relatedness):
   Look for signs of:
   - Competence: positive performance indicators
     (high kill count, achievements, skill displays),
     successful actions
   - Autonomy: varied equipment/loadout, player-
     chosen paths, customized appearance
   - Relatedness: team coordination visible, group
     activity, communication indicators
   -> Low engagement signs: poor performance shown,
      default/basic equipment, isolated player

4. MDA Framework (challenge, sensation, fellowship,
   narrative):
   Look for signs of:
   - Challenge: competitive situation, close match,
     difficult enemies, survival pressure
   - Sensation: visual effects (explosions,
     particles, lighting), dynamic camera, action-
     heavy scene
   - Fellowship: multiple players engaged together,
     team-based activity
   - Narrative: story moments, dramatic events,
     critical match situations (final round,
     overtime)
   -> Low engagement signs: no visual excitement,
      static scene, no narrative tension

DECISION GUIDELINES:
- Count how many theories suggest High vs Low
  engagement
- If the frame shows a menu, pause screen, or
  loading screen -> almost always Low
- When signals conflict (e.g., active scene but
  poor performance), lean toward the majority of
  indicators

OUTPUT (strict format):
Label: [High/Low]
Rationale: [One sentence: describe the key visual
features you observed and which theory dimensions
they satisfy or violate]
\end{lstlisting}
\normalsize

\paragraph{4-Shot Demonstration Examples.}
Each example includes a DOOM frame paired with a gold label and a theory-grounded rationale:

\begin{itemize}[nosep,leftmargin=*]
    \item \textbf{Example 1 (High):} Low HP clutch with C4 planted and timer under 30s $\Rightarrow$ active challenge (Flow), high control (GameFlow), competence challenge (SDT), narrative tension (MDA).
    \item \textbf{Example 2 (Low):} Full HP, empty kill feed, no objectives $\Rightarrow$ no feedback (Flow), low immersion (GameFlow), no competence indicators (SDT), no aesthetic engagement (MDA).
    \item \textbf{Example 3 (High):} Active kill feed with headshot, 4 teammates visible, MVP display $\Rightarrow$ positive feedback (Flow), social immersion (GameFlow), competence satisfaction (SDT), sensation + fellowship (MDA).
    \item \textbf{Example 4 (Low):} Pause menu visible $\Rightarrow$ disrupted flow (Flow), broken immersion (GameFlow), no competence demonstration (SDT), eliminated challenge/narrative (MDA).
\end{itemize}

\section{Demonstration Order Experiment}
\label{app:order}

We test whether the ordering of 4-shot examples affects predictions (Table~\ref{tab:order}).
All orderings produce nearly identical distributions, confirming that the model's over-prediction of High is an intrinsic bias, not a positional artifact.

\begin{table}[ht]
\centering
\small
\begin{tabular}{lcc}
\toprule
\textbf{Ordering} & \textbf{Pred.\ High} & \textbf{Pred.\ Low} \\
\midrule
Original [L, H, H, L] & 12 & 8 \\
Last-high [L, L, H, H] & 12 & 8 \\
Alternating H-L & 10 & 10 \\
Alternating L-H & 12 & 8 \\
\bottomrule
\end{tabular}
\caption{Prediction distribution under different 4-shot example orderings (CS:GO Office subset). Distributions are stable across orderings.}
\label{tab:order}
\end{table}

A chi-square test of independence confirms no significant association between example ordering and prediction distribution ($\chi^2 = 0.614$, $df = 3$, $p = 0.893$), indicating that the over-prediction of High is an intrinsic model bias rather than a positional artifact of the demonstration examples.

\section{Confidence--Accuracy Analysis}
\label{app:error_viz}

We estimate model confidence from the generated rationale text:
\begin{equation}
C(r) = \frac{|\{w \in r : w \in \mathcal{A}\}|}{|\{w \in r : w \in \mathcal{A} \cup \mathcal{H}\}|}
\end{equation}
where $\mathcal{A}$ is a set of assertive words (``clearly,'' ``demonstrates,'' ``confirms'') and $\mathcal{H}$ is a set of hedging words (``possibly,'' ``minimal,'' ``lacking'').
The mean confidence for incorrect predictions (0.597) is higher than for correct predictions (0.453).

\begin{figure}[t]
\centering
\begin{tikzpicture}
\begin{axis}[
    ybar,
    bar width=20pt,
    width=\columnwidth,
    height=5.5cm,
    ylabel={Accuracy (\%)},
    xlabel={Confidence Level},
    symbolic x coords={High, Medium, Low},
    xtick=data,
    x tick label style={font=\small},
    ymin=0, ymax=80,
    ymajorgrids=true,
    grid style=dashed,
    nodes near coords={\pgfmathprintnumber\pgfplotspointmeta\%},
    nodes near coords style={font=\small},
    every node near coord/.append style={yshift=3pt},
    point meta=explicit,
]
\addplot[fill=orange!70, draw=orange!90] coordinates {(High,38.5) [38.5] (Medium,25.0) [25.0] (Low,64.3) [64.3]};
\node[font=\tiny] at (axis cs:High,6) {$n=26$};
\node[font=\tiny] at (axis cs:Medium,6) {$n=64$};
\node[font=\tiny] at (axis cs:Low,6) {$n=28$};
\draw[dashed, gray, thick] ({rel axis cs:0,0}|-{axis cs:High,52.4}) node[above right, font=\tiny] {Random} -- ({rel axis cs:1,0}|-{axis cs:High,52.4});
\end{axis}
\end{tikzpicture}
\caption{Prediction accuracy by VLM confidence level, showing severe miscalibration. High-confidence predictions are the least accurate (38.5\%), while low-confidence predictions are the most accurate (64.3\%). The dashed line shows random baseline (52.4\%).}
\label{fig:confidence_accuracy}
\end{figure}
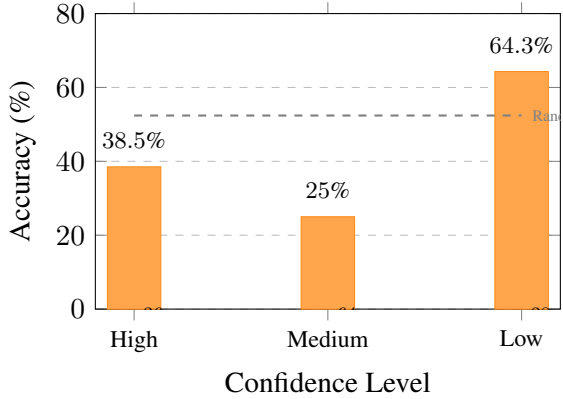

\section{Qualitative Error Examples}
\label{app:errors}

The following examples are drawn from real VLM errors in our theory-guided zero-shot (S2) evaluation on CSGO18 and CS:GO Office, organized by failure mode. Each case shows the VLM's actual rationale and the corresponding ground-truth annotation.

\paragraph{Example 1: Static scene misclassification (Type~A).}

\begin{figure}[ht]
\centering
\includegraphics[width=0.85\columnwidth]{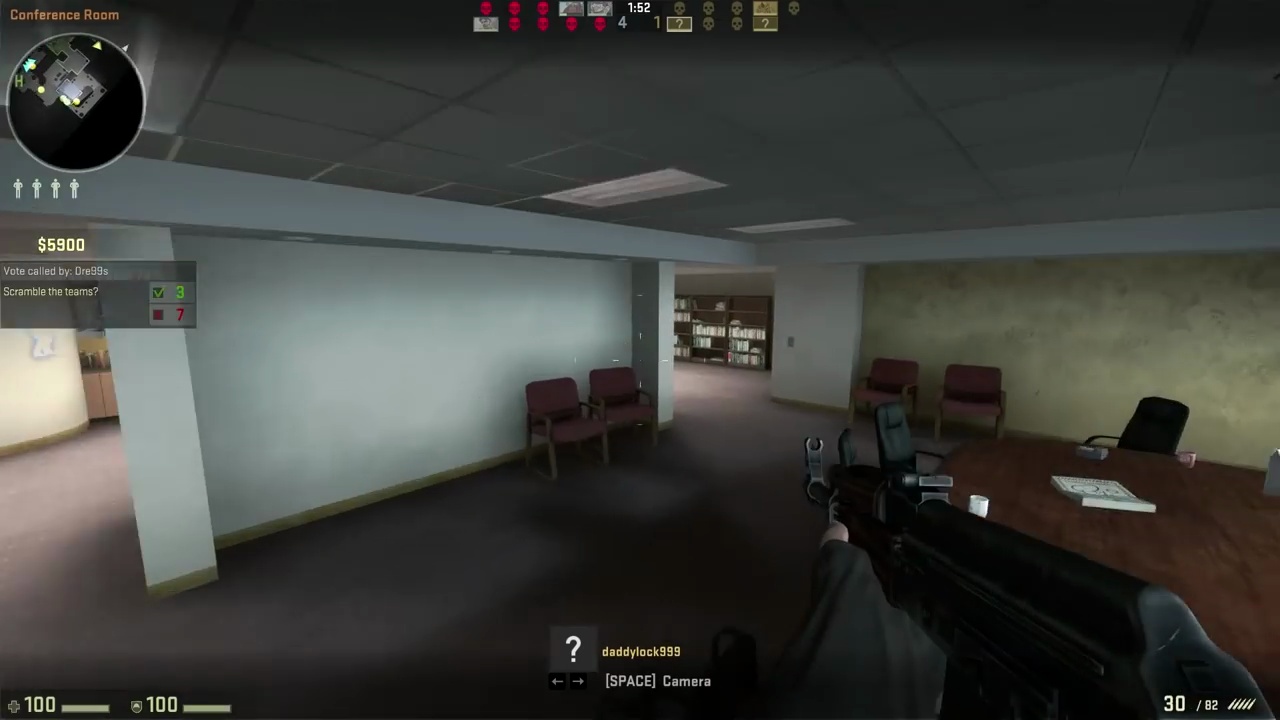}
\caption{CS:GO Office, frame 24--25. The VLM predicts Low due to the static vote screen, but the ground truth is High.}
\label{fig:error_ex1}
\end{figure}

\noindent\textit{Ground truth:} High. \quad \textit{VLM prediction:} Low. \\
\textit{VLM rationale:} ``The scene shows a player in a conference room with no immediate action or visible enemies, and a vote is occurring, indicating a pause in gameplay. Feedback elements like score changes are present, but there's no active challenge or player control evident, suggesting low engagement.'' \\
\textit{Analysis:} The VLM treats the vote screen as a ``pause,'' failing to infer that an ongoing team vote during a close match represents a moment of high strategic engagement. This exemplifies FM3 (Post-Match Context Blindness, Section~\ref{sec:fm3}): the model equates visual stillness with low engagement.

\paragraph{Example 2: Sensation shortcut (Type~C).}

\begin{figure}[ht]
\centering
\includegraphics[width=0.85\columnwidth]{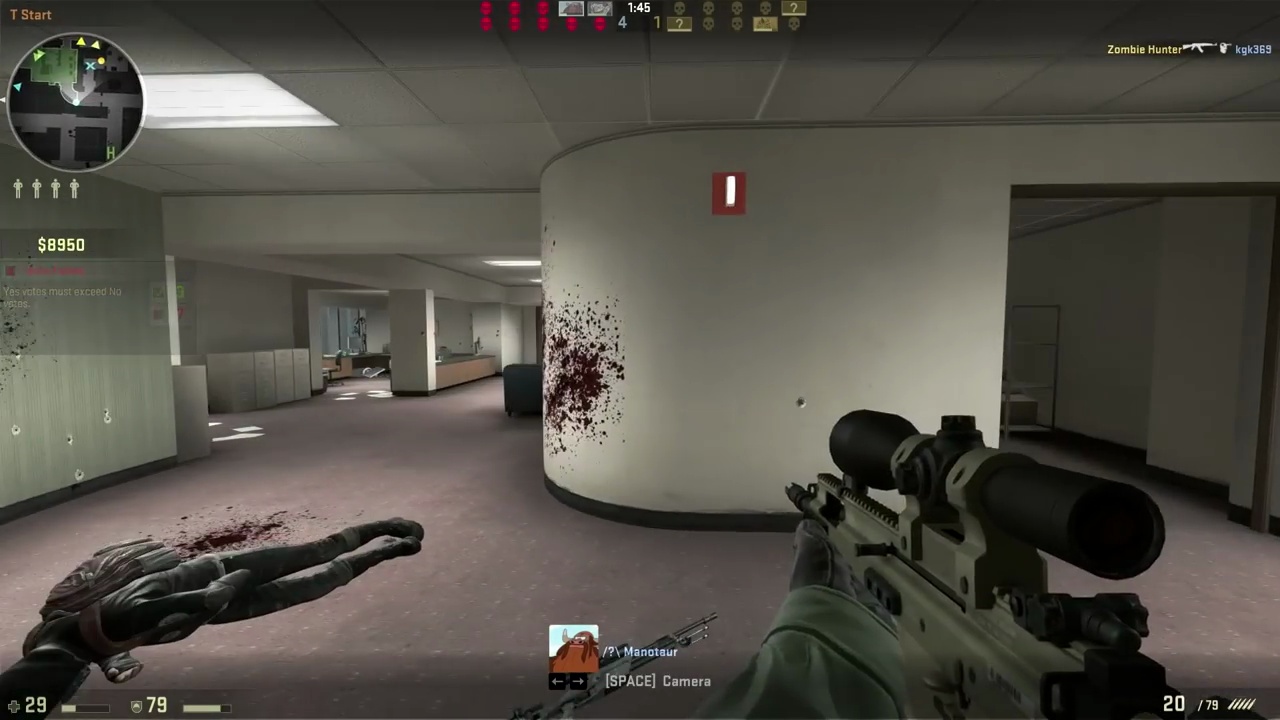}
\caption{CS:GO Office, frame 31--32. Blood splatters and weapon animations trigger a High prediction, but the ground truth is Low.}
\label{fig:error_ex2}
\end{figure}

\noindent\textit{Ground truth:} Low. \quad \textit{VLM prediction:} High. \\
\textit{VLM rationale:} ``The frame shows active combat with visible feedback (blood splatter, weapon in use), clear objectives (mini-map, score), and player control (aiming), indicating engagement across multiple theories.'' \\
\textit{Analysis:} Blood splatters and weapon animations trigger a High prediction despite the ground truth being Low. The VLM conflates visual intensity with engagement, illustrating FM1 (Visual Intensity Bias, Section~\ref{sec:fm1}).

\paragraph{Example 3: Relatedness shortcut (Type~B).}

\begin{figure}[ht]
\centering
\includegraphics[width=0.85\columnwidth]{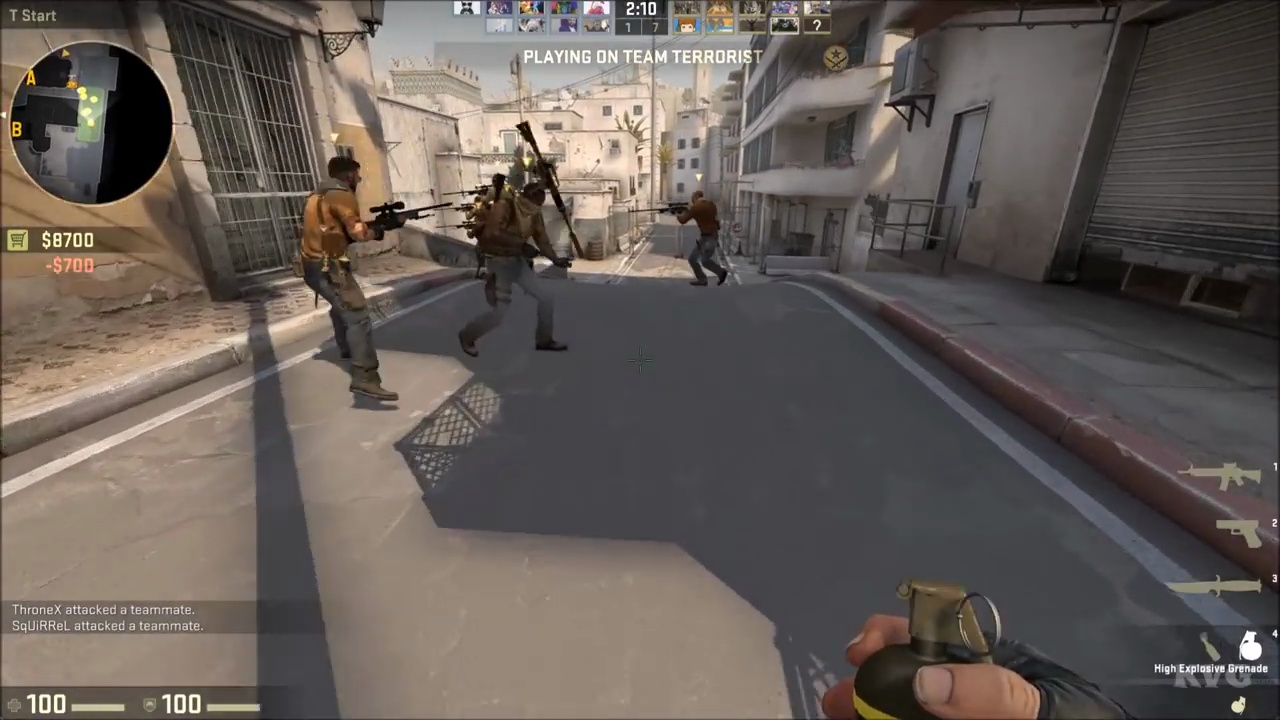}
\caption{CSGO18, frame 15--16. Visible teammates and HUD indicators lead to a High prediction, but the ground truth is Low.}
\label{fig:error_ex3}
\end{figure}

\noindent\textit{Ground truth:} Low. \quad \textit{VLM prediction:} High. \\
\textit{VLM rationale:} ``The frame shows active combat with multiple players, visible health and score indicators, and immediate feedback from teammate actions, satisfying challenge, feedback, and social presence dimensions of engagement theories.'' \\
\textit{Analysis:} The mere visibility of teammates and HUD indicators (health bars, scores) triggers a High prediction. The VLM uses ``social presence'' as a shortcut for engagement without evaluating whether the moment is genuinely engaging, exemplifying FM2 (Surface Feature Shortcuts, Section~\ref{sec:fm2}).

\paragraph{Example 4: Challenge misassessment (Type~D).}

\begin{figure}[ht]
\centering
\includegraphics[width=0.85\columnwidth]{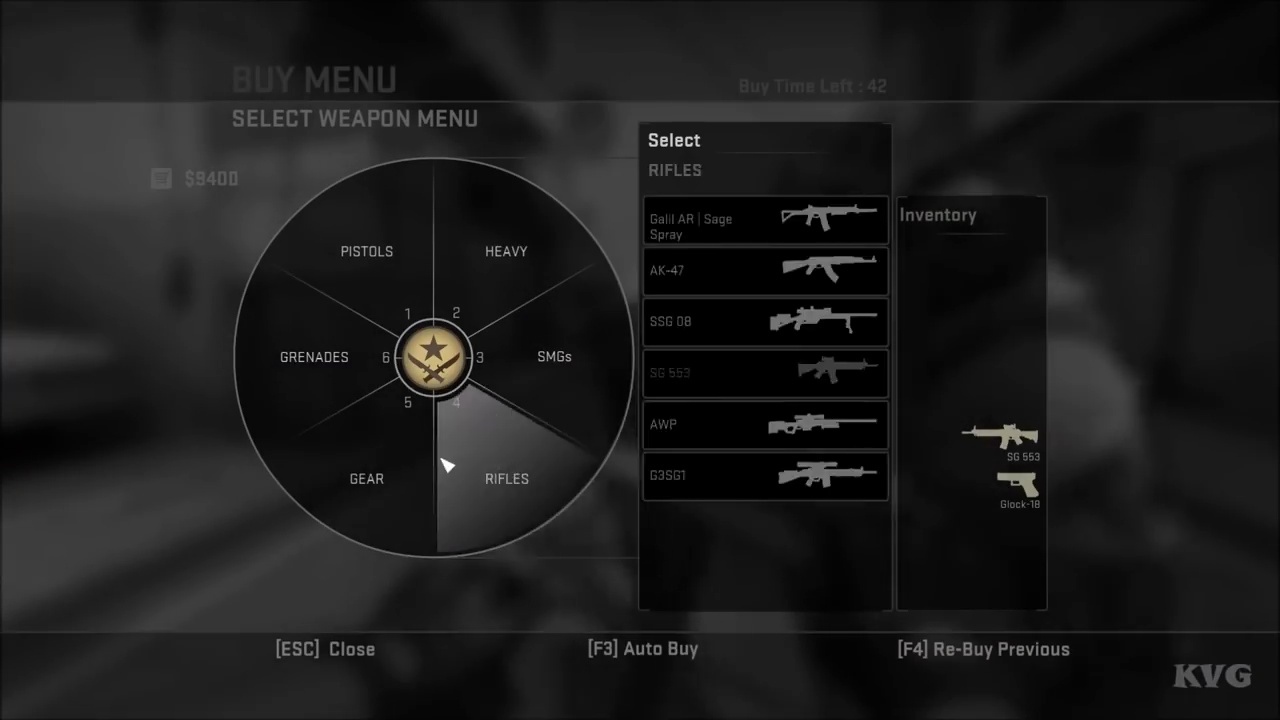}
\caption{CSGO18, frame 12--13. The buy menu is dismissed as ``no active gameplay,'' but the ground truth is High.}
\label{fig:error_ex4}
\end{figure}

\noindent\textit{Ground truth:} High. \quad \textit{VLM prediction:} Low. \\
\textit{VLM rationale:} ``The frame shows a buy menu with no active gameplay, indicating a pause in action. This suggests a lack of immediate challenge, goals, and feedback, aligning with low engagement according to Flow Theory and GameFlow.'' \\
\textit{Analysis:} The buy menu is dismissed as ``no active gameplay,'' yet this phase---selecting equipment under time pressure between rounds---is annotated as High engagement. The VLM fails to recognize that strategic decision-making constitutes engagement, instead requiring visible combat as evidence of challenge.

\end{document}